\documentclass{article}

\usepackage[eandd,final,nonatbib]{neurips_2026}

\usepackage[utf8]{inputenc}
\usepackage[T1]{fontenc}
\usepackage{xcolor}
\usepackage{xurl}
\usepackage{url}
\usepackage[hidelinks]{hyperref}
\usepackage{booktabs}
\usepackage{array}
\usepackage{amsfonts}
\usepackage{amsmath}
\usepackage{amssymb}
\usepackage{comment}
\usepackage{nicefrac}
\usepackage{microtype}
\usepackage{xcolor}
\usepackage{graphicx}
\usepackage{float}
\usepackage{subcaption}
\usepackage[
  backend=biber,
  style=authoryear,
  natbib=true,
  giveninits=true,
  maxcitenames=1,
  maxbibnames=99,
  dashed=false,
  uniquename=false,
  uniquelist=false,
  labeldateparts=true
]{biblatex}

\title{\textsc{Chess-World-Model}: A $10$M-Game Benchmark for Exact State Tracking from Chess Move Sequences}

\author{
    Benjamin Walker \\
    Mathematical Institute, University of Oxford \\
    Continuous Intelligence Limited \\
    \texttt{contact@benwalker.co.uk}
    \And
    Terry Lyons \\
    Mathematical Institute, University of Oxford \\
    Department of Mathematics, Imperial College London
    }

\begin{document}

\maketitle

\begin{abstract}
World models require state tracking, which is the ability to maintain a correct latent state across action sequences. Existing benchmarks are often synthetic or language-based, limiting their value as tests of structured state updates in realistic domains. We introduce \textsc{Chess-World-Model}, a large-scale state-tracking benchmark built from $10$ million real chess games, where models predict the exact board state reached after a sequence of legal moves. Alongside a held-out real-game split, we include an out-of-distribution test set of uniformly random legal play both at its original trajectory lengths and after matching the human-game length distribution. This tests whether models learn the transition rules rather than shortcuts from common human positions. Prior theoretical and empirical work has shown that Transformers struggle to state-track, while input-dependent linear RNNs require expressive state-transition matrices to do so. We therefore benchmark a causal Transformer, block-diagonal SLiCE, Mamba-3, and Gated DeltaNet with negative eigenvalues under a matched interface and training protocol. The recurrent models strongly outperform the Transformer at approximately $3$ and $8$ million parameters. Real-game performance largely saturates by the $18$ million parameter scale, but length-matched random evaluation continues to distinguish architectures through the largest tested scale of approximately $43$ million parameters. Additionally, ablations show that less expressive state-transition mechanisms generally reduce performance on the out-of-distribution splits. Together, these results establish \textsc{Chess-World-Model} as a practical large-scale benchmark for state tracking that exposes failures model scale would otherwise conceal.
\end{abstract}

\section{Introduction}
\label{sec_intro}

\subsection{Motivation}

A world model is only as useful as the state it maintains.
As actions and observations accumulate, the model must update this state, preserve information over long horizons, and remain consistent when trajectories depart from familiar training patterns.
This requirement appears in classical model-based reinforcement learning and predictive state representations \citep{sutton1990dyna,littman2001predictive}, in neural world models that learn latent dynamics for control \citep{ha2018world,hafner2020dreamer}, and in recent joint-embedding methods that learn representations by predicting images and videos in latent space \citep{lecun2022path,assran2023ijepa,assran2025vjepa2}.
In all of these settings, prediction is not merely a one-step pattern-matching problem.
It depends on whether the model can maintain and update a coherent representation of the world.

Despite this central role, exact state maintenance remains difficult to evaluate.
Many world-model tasks couple state tracking to perception, reward prediction, planning, or language understanding, making failures hard to localise.
Conversely, many formal state-tracking benchmarks isolate the computation cleanly but rely on synthetic tasks such as parity, modular arithmetic, regular languages, or permutation composition \citep{deletang2023neural,cirone2024theoretical,grazzi2025unlocking,merrill2024illusion}.
These tasks have been crucial for exposing architectural limitations.
However, because they are often small or highly regular, increasing model scale can obscure state-tracking failures by allowing models to exploit task-specific shortcuts or memorise large parts of the problem structure \citep{deletang2023neural,li2025how}.
They therefore leave open a complementary empirical question.
Can modern sequence models maintain a structured latent state in a realistic deterministic action domain?

Chess provides a natural testbed for this question.
The task is not to play chess, but to act as a learned game engine.
Given only a sequence of legal moves, the model must reconstruct the exact state reached by the game.
This state includes both the board configuration and the auxiliary variables needed to specify the position.
The transition rules are known exactly, and every prediction can be checked against the rules of chess without learned simulators or subjective judgments.

Chess also provides a principled axis of distribution shift.
Human games occupy a structured and strategically meaningful subset of legal play, just as observations of the physical world occupy only a small subset of all dynamically possible trajectories.
However, a world model should not merely interpolate within familiar trajectories.
It should infer the underlying dynamics well enough to remain consistent under distribution change.
Uniformly random legal play tests this distinction directly.
It produces positions that are valid under the same transition rules, but that depart sharply from human regularities.
Strong performance on held-out human games may therefore reflect memorised openings, common tactical motifs, or likely piece configurations.
Strong performance under random legal play requires robust rule-consistent state updates.

\subsection{Contributions}

We introduce \textsc{Chess-World-Model}, a large-scale benchmark for exact state tracking from chess move sequences.
The benchmark is constructed from $10$M real games.
Given a sequence of moves, models must predict the corresponding sequence of game states, including piece locations and auxiliary state variables.
This sequence-to-sequence formulation allows every prediction to be evaluated exactly.
Alongside held-out real games, we evaluate uniformly random legal play both at its original trajectory lengths and after matching the human-game length distribution.
The length-matched evaluation tests state tracking beyond familiar human trajectories at the same horizons, while the original random split additionally tests generalisation to the longer trajectories produced by random play.

Recent theoretical and empirical work has shown that Transformers and input-dependent linear RNNs with diagonal state-transition matrices, such as Mamba and Mamba-2, struggle with state-tracking tasks \citep{merrill2022saturated,merrill2024illusion,walker2025slice}.
By contrast, richer recurrent transition mechanisms, such as input-dependent state-transition matrices with negative eigenvalues, block-diagonal structure, or diagonal-plus-low-rank structure, are theoretically more expressive and empirically improve state-tracking performance \citep{grazzi2025unlocking, fan-etal-2024-advancing,walker2025slice,siems2025deltaproductimprovingstatetrackinglinear,lahoti2026mamba3}.

Motivated by these results, we benchmark a causal Transformer against three recent input-dependent linear RNNs with expressive state transitions: block-diagonal SLiCE, Mamba-3, and Gated DeltaNet with negative eigenvalues \citep{walker2025slice, grazzi2025unlocking, lahoti2026mamba3}.
All models are trained under a shared prediction interface and matched training protocol across four architecture scales, $\mathrm{d}128/\mathrm{l}1$, $\mathrm{d}256/\mathrm{l}2$, $\mathrm{d}384/\mathrm{l}4$, and $\mathrm{d}512/\mathrm{l}6$, where $\mathrm{d}$ denotes hidden dimension and $\mathrm{l}$ denotes number of layers.
These correspond to approximately $3$, $8$, $18$, and $40$ million parameters, respectively.

Our experiments yield three main findings.
First, the recurrent models substantially outperform the Transformer at the two smaller scales, $\mathrm{d}128/\mathrm{l}1$ and $\mathrm{d}256/\mathrm{l}2$.
Second, held-out real-game performance largely saturates by $\mathrm{d}384/\mathrm{l}4$, whereas length-matched random evaluation remains discriminative through the largest tested scale, $\mathrm{d}512/\mathrm{l}6$.
The original random split additionally exposes longer-horizon failures that can change the relative performance of the architectures.
Third, ablations generally favour more expressive recurrent state-transition mechanisms under distribution shift.
Together, these results establish \textsc{Chess-World-Model} as a practical large-scale benchmark for state tracking that exposes failures hidden by scale and in-distribution evaluation.
It provides a foundation for developing expressive and efficient sequence models whose update mechanisms support reliable state tracking, a necessary ingredient for learned world models.

Code for constructing \textsc{Chess-World-Model} from the public Lichess Open Database, generating the random-uniform test split, training the evaluated models, and computing the reported metrics is available at \url{https://github.com/Benjamin-Walker/Chess-World-Model}. The benchmark data, including all three test splits, are available at \url{https://doi.org/10.7910/DVN/C8TEVB}.

%We provide the dataset through an anonymous Harvard Dataverse link,
%\url{https://dataverse.harvard.edu/previewurl.xhtml?token=64ee751f-0a75-48db-8ce5-9970a1fa17bf},
%and the code through an anonymous repository, 
%\url{https://anonymous.4open.science/r/world-modelling-anon-C787}.
%The full dataset is $8.82$GB compressed and $116.40$GB uncompressed. 
%The held-out real-game test set is $119$MB, and downloadable separately as a lightweight preview of the data.
%Upon acceptance, the dataset will be released under CC0 and the code under the MIT License.
%Both the dataset and code will be released publicly upon acceptance.

\subsection{Related work}

Chess is a natural domain for studying state tracking in sequence models, since move sequences induce an exactly defined latent board state.
\citet{toshniwal2022chess} introduced chess as a testbed for language-model state tracking, training Transformer language models to predict move sequences and evaluating legal move prediction together with probes of the internal board state.
Related interpretability work has studied whether chess models trained on game transcripts learn internal representations of board state and player skill \citep{karvonen2024emergent}.
\citet{cooper2025pgn2fen} introduced the PGN-to-FEN benchmark, which tests whether pretrained large language models can reconstruct the exact final board state from a move sequence.
\citet{harang2025tracking} introduced a closely related state-affordance evaluation that also prompts language models to infer chess states, but scores predictions by comparing the legal continuations induced by the predicted and true positions.
On the theoretical side, \citet{merrill2024illusion} formalised chess in source-target move notation as an example of state tracking.

More broadly, deterministic board games have been used to study whether sequence models learn implicit world models.
In Othello, models trained on move sequences have been shown to contain internal representations aligned with the underlying board state \citep{li2023othello,nanda2023linear}.
At the same time, recent work argues that strong predictive performance can coexist with incoherent or inconsistent recovered world states, motivating evaluations that directly test state recovery rather than only next-action prediction \citep{vafa2024evaluating}.
This motivates using chess not only as a language-modelling domain, but as an exactly checkable state-tracking environment.

\textsc{Chess-World-Model} builds on this line of work by turning chess state tracking into a large-scale supervised benchmark for sequence-model architectures.
It closely instantiates the formal source-target chess state-tracking problem studied by \citet{merrill2024illusion}, while extending the target state to include auxiliary FEN-style variables.
It differs from prior chess-based empirical evaluations in three main respects.
First, it trains models to predict the complete game state at every prefix of a move sequence, rather than using next-move training with state probes or prompting pretrained language models for a final position \citep{toshniwal2022chess,cooper2025pgn2fen,harang2025tracking}.
Second, it evaluates exact sequence-to-sequence state reconstruction, rather than only final-position recovery or indirect consistency through legal continuations.
Third, it includes an out-of-distribution random-agent split, generated by uniformly random legal play, which tests whether models learn rule-consistent state evolution beyond the distribution of human games.
Together, these choices make \textsc{Chess-World-Model} suitable for comparing sequence-model architectures as candidate state-update mechanisms for learned world models.

\section{\textsc{Chess-World-Model}}

\subsection{Chess as a deterministic state-tracking environment}
\label{subsec:chess-state-tracking}

\begin{figure}[t]
    \centering

    \begin{subfigure}{0.46\textwidth}
        \centering
        \includegraphics[width=\textwidth]{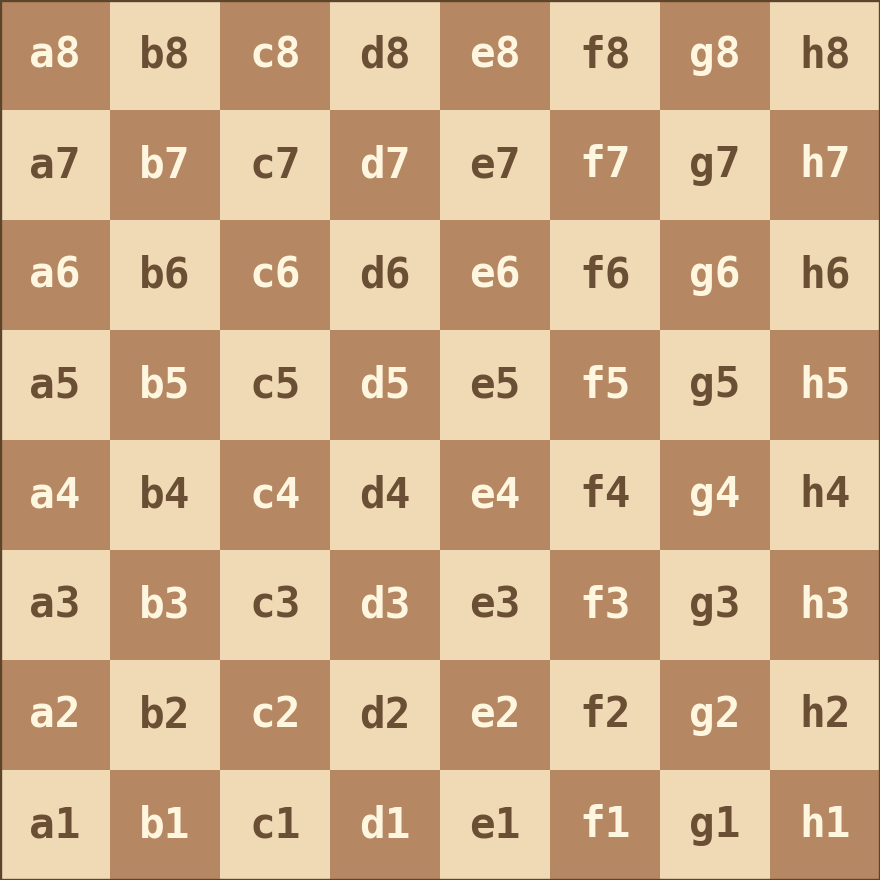}
        \caption{The standard algebraic coordinates for board squares. Columns, called files, are labelled $a,\ldots,h$, and rows, called ranks, are labelled $1,\ldots,8$.}
        \label{fig:chess-coordinate-system}
    \end{subfigure}
    \hfill
    \begin{subfigure}{0.46\textwidth}
        \centering
        \includegraphics[width=\textwidth]{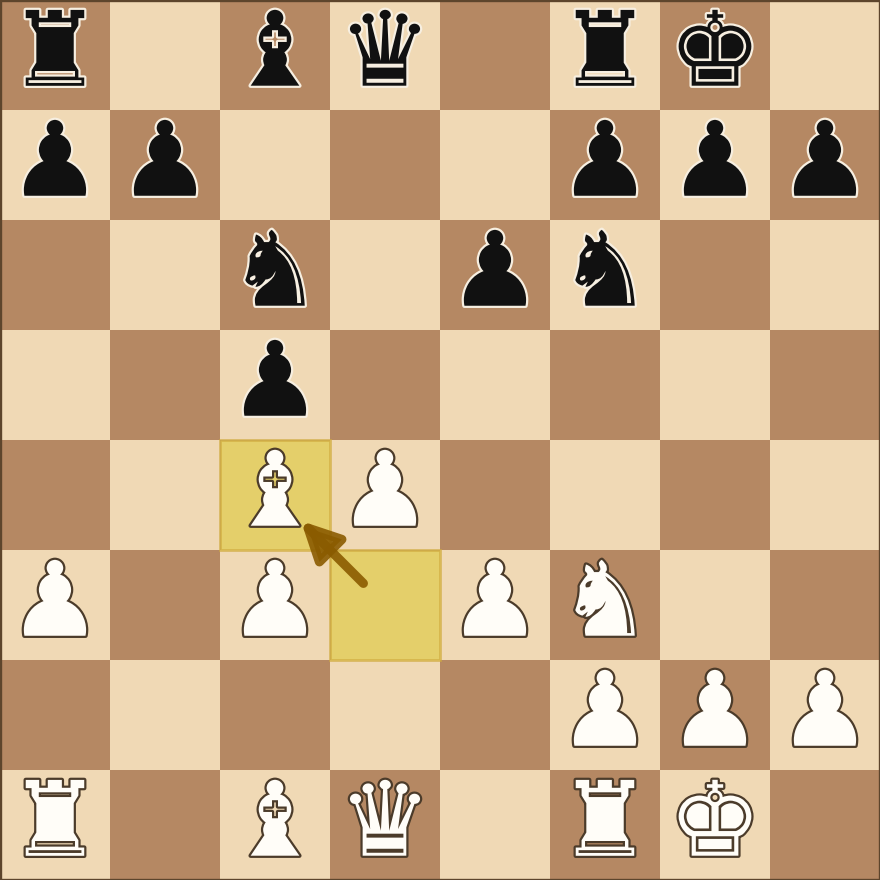}
        \caption{A move in UCI notation is represented by concatenating its source and target squares. Here the highlighted move is $\mathrm{d3c4}$.}
        \label{fig:chess-uci-move-notation}
    \end{subfigure}

    \caption{Chess move notation used in our dataset. The left panel shows the board coordinates used to identify squares. The right panel illustrates UCI move notation \citep{meyerkahlen2000uci}, in which each move is recorded by its source square followed by its target square.}
    \label{fig:chess-move-notation-overview}
\end{figure}

Standard chess is a simple world. The initial state is fixed, the dynamics are deterministic, and the action space is fully observed, regular, and noiseless. Unlike domains with perceptual ambiguity, stochastic transitions, or hidden exogenous variables, chess allows us to study state tracking in a setting where uncertainty about the world comes primarily from the need to integrate an action history correctly. This makes the domain useful for isolating the memory and update mechanisms of sequence models from confounds introduced by sensing, control, or environment noise.

We use the standard rules of chess as specified by the FIDE Laws of Chess \citep{fide2023laws}, and encode moves using UCI notation \citep{meyerkahlen2000uci}. As shown in Figure~\ref{fig:chess-move-notation-overview}, board squares are indexed by standard algebraic coordinates. Columns, called files, are labelled $a,\ldots,h$, and rows, called ranks, are labelled $1,\ldots,8$. A move is encoded by concatenating its source and target squares. For example, the move $\mathrm{d3c4}$ denotes a piece moving from square $d3$ to square $c4$. Promotion moves are encoded by appending the promoted piece type, for example $\mathrm{e7e8q}$ for promotion to a queen.

Although each move is directly observed, the game state is not provided to the model. It must be reconstructed from the move stream alone. This requires more than remembering the current piece placement. In this paper, we use game state to mean the full FEN-style chess state reached after a move prefix \citep{edwards1994pgn}, which includes whose turn it is, which castling rights remain available, whether an en passant capture is currently legal, the halfmove clock, and the fullmove number, but excludes the history of previous board states needed to adjudicate repetition-based draws. These quantities are part of the represented state because they affect legal continuations, draw-related counters, or the formal description of the position. The complete decomposition of this represented state into categorical prediction targets is given in Appendix~\ref{app-state-label-layout}.

\subsection{Task and trajectory alignment}
\label{subsec:task-trajectory-alignment}

The benchmark is a move-to-state trajectory prediction task rather than a terminal-board reconstruction task. Each example is a legal chess trajectory beginning from the standard initial position. The input is the move sequence, while the target is the complete chess state after each observed prefix of that sequence.

\begin{figure}[t]
    \centering
    \includegraphics[width=0.9\textwidth]{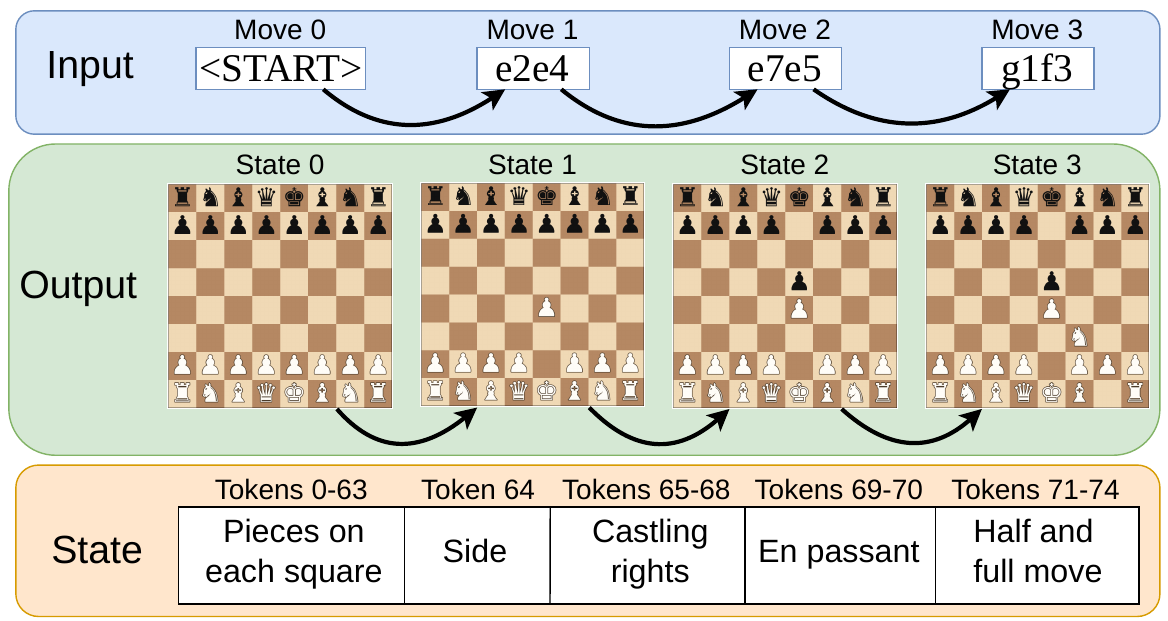}
    \caption{\textsc{Chess-World-Model} as an aligned move-to-state prediction task. The input sequence consists of a start token followed by UCI moves. Each input prefix is aligned with the complete chess state obtained after applying that prefix. The target state contains the piece occupying each square, the side to move, castling rights, en passant information, and the halfmove and fullmove counters.}
    \label{fig:chess-dataset-overview}
\end{figure}

Figure~\ref{fig:chess-dataset-overview} illustrates the alignment. The first input token is a start symbol, aligned with the initial state. Each subsequent input token is the categorical encoding of a UCI move, aligned with the state obtained after applying that move. Thus, if the observed plies are $m_{1:T}$, the data pipeline constructs an aligned sequence of move tokens $x_{0:T}$ and chess states $s_{0:T}$ such that
\[
x_0 = \texttt{<START>}, \qquad s_0 = s_{\mathrm{init}},
\]
and for each $t \geq 1$,
\[
x_t = \mathrm{enc}_{\mathrm{move}}(m_t), \qquad s_t = \mathrm{State}(m_{1:t}).
\]
Here $\mathrm{enc}_{\mathrm{move}}$ is the fixed move encoding map described in Section~\ref{subsec:move-state-encoding}: it maps each UCI move to a single categorical token by its source square, target square, and promotion type. The model receives the encoded move prefix $x_{0:t}$ and predicts the full chess state $s_t$ at every timestep. Every game therefore contributes supervision at every observed prefix length, including the initial position. Appendix~\ref{app-data-preprocessing} provides additional details on PGN replay, trajectory construction, filtering, and deterministic evaluation.

\subsection{Move and state encoding}
\label{subsec:move-state-encoding}

Moves are represented with a fixed packed vocabulary over $64 \times 64 \times 5 = 20{,}480$ move geometries. The first two factors encode the source and target squares, and the final factor encodes the promotion type. The promotion value is one of no promotion, queen, rook, bishop, or knight. The implementation appends two special symbols, \texttt{<START>} and \texttt{<PAD>}, giving a total move vocabulary size of $20{,}482$. This encoding does not attempt to enumerate only legal moves. Legality is enforced by the source trajectories. Appendix~\ref{app-move-token-coverage} reports move-token coverage and verifies that every packed move id appearing in the random-uniform test set is observed in the training data.

Each target state is represented as $75$ categorical labels. The first $64$ labels encode the board in square order $a8,b8,\ldots,h1$, using the piece-state vocabulary $\mathcal{P}
=
\{\text{empty}\}
\cup
\{ \text{white } p, \text{black } p \mid p \in \{\text{pawn, knight, bishop, rook, queen, king}\}\}.$ The remaining $11$ labels encode the FEN-style auxiliary state reached after the move prefix \citep{edwards1994pgn}, excluding the history of previous board states needed to adjudicate repetition-based draws. These labels encode the side to move, castling rights, en passant availability, the halfmove clock, and the fullmove number. The full per-label layout is given in Appendix~\ref{app-state-label-layout}, and the same grouping is shown schematically in Figure~\ref{fig:chess-dataset-overview}.

\subsection{Exactness metrics}
\label{subsec:exactness-metrics}

The main body of the paper focuses on exact state accuracy, since the benchmark is designed to test whether a model recovers the complete chess state induced by a move prefix. Let $J$ denote the set of $75$ state targets, and write $\hat{s}_{t,j}$ for the prediction of target $j$ at timestep $t$. The timestep-level exact state metric is $\mathrm{ExactState}(t)
=
\mathbf{1}\{\hat{s}_t = s_t\}.$ This score is equal to $1$ only when all $75$ state labels are predicted correctly. A single error in piece placement, side to move, castling rights, en passant state, or the move counters makes the state incorrect. Appendix~\ref{app-data-preprocessing} defines additional metrics for partial state accuracy, trajectory exactness, and length-binned prefix exactness, with the corresponding results reported in Appendices~\ref{app-test-results}, \ref{app-variant-results}, and \ref{app-temporal-bin-results}.

\subsection{Data source and split construction}
\label{subsec:data-source-splits}

The real-game corpus is derived from the public Lichess open database, whose monthly PGN exports are released under a CC0 license \citep{lichess2026database}. During preprocessing, each retained game is converted into the aligned move-state trajectory described in Section~\ref{subsec:task-trajectory-alignment}. The processed examples also store lightweight bookkeeping fields, such as game identifier, result, and length. Usernames, clock comments, annotations, and other metadata that are unnecessary for the benchmark are not used as modelling inputs. Games shorter than $10$ full moves are filtered out.

The current benchmark instantiation uses $10$M real games as the training  and validation source. Training and validation are game-disjoint. Rather than materialising a separate validation file, the implementation assigns each game to train or validation by hashing its game identifier into a fixed bucket set. With the default settings, approximately $0.5\%$ of games are reserved for validation. This gives a deterministic split that is reproducible across runs.

Final evaluation uses three fixed test conditions, each containing $10{,}000$ trajectories.
The first  is a held-out real-game split from the same overall source family. This measures generalisation within the distribution of human play. The second is generated by sampling uniformly from the legal move set at each ply until game termination, then retaining $10$k games of length at least $10$ full moves. This measures robustness under legal but non-human action distributions. 
The third is a length-matched random split, constructed by truncating distinct random legal trajectories to the lengths of the human test games.
Throughout, random-uniform refers to the original random split, while length-matched random refers to this additional evaluation.
Appendix~\ref{app-length-matched} describes the construction, including its conditioning on trajectories reaching the required length.
The data splits are summarised in Table~\ref{tab:benchmark-splits}. Further details on preprocessing, the hash-based validation split, random-uniform generation, and evaluation protocol are given in Appendix~\ref{app-data-preprocessing}.

\begin{table}[t]
\centering
\small
\begin{tabular}{lll}
\toprule
Split & Size & Description \\
\midrule
Real-game train and validation & $10$M source games & Processed Lichess games \\
Held-out real validation & $\approx 0.5\%$ by default & Deterministic game split used for model selection \\
Held-out real test & $10$k & Fixed real-game benchmark set \\
Random-uniform test & $10$k & Fixed legal self-play benchmark set \\
Length-matched random test & $10$k & Random play at human-game lengths \\
\bottomrule
\end{tabular}
\caption{Benchmark splits in the current implementation. The main distribution shift is from human play to uniformly random legal self-play, while keeping the same chess dynamics and target representation.}
\label{tab:benchmark-splits}
\end{table}

\section{Experimental methodology}
\label{sec:exp}

\subsection{Shared prediction interface}

All architectures are evaluated through the same move-to-state prediction interface. Each model receives the packed move sequence $x_{0:T}$ and predicts the aligned chess state $s_{0:T}$ at every timestep. The output layer is factorised into target-specific categorical heads. It consists of $64$ board heads with $13$ classes each, one binary side-to-move head, four binary castling heads, one $9$-way en passant file head, one $3$-way en passant rank head, and two $256$-way byte heads for each of the halfmove and fullmove counters.

For a trajectory $(x_{0:T},s_{0:T})$, the training objective is the masked sum of per-target cross-entropies over all non-padded timesteps,
\[
\mathcal{L}(x_{0:T}, s_{0:T})
=
-\frac{1}{|J|(T+1)}\sum_{t=0}^{T}\sum_{j \in J}
\log p_\theta(s_{t,j} \mid x_{0:t}),
\]
where $J$ denotes the set of $75$ state targets. This keeps the supervision signal, output representation, and evaluation metrics fixed across model families. Differences in performance therefore reflect how each architecture processes the move history, rather than differences in the prediction task.

The main architectural distinction is between attention over the whole observed prefix and recurrent state updates. A causal Transformer can revisit earlier moves through self-attention at every timestep. The recurrent models instead compress the observed history into a hidden state that is updated online as new move tokens arrive. Chess is useful for this comparison because the input stream is simple, but the latent state that must be maintained is exact, structured, and history-dependent.

\subsection{Architectures}

We compare four sequence-model families. The first is a causal Transformer \citep{vaswani2017attention}, which serves as an attention-based baseline. The remaining three are recurrent architectures with input-dependent state updates: block-diagonal SLiCE \citep{walker2025slice}, Mamba-3 \citep{gu2024mamba,Dao2024Transformers,lahoti2026mamba3}, and Gated DeltaNet \citep{yang2025gateddeltanetworksimproving}. For Gated DeltaNet, the main configuration allows negative eigenvalues, following recent evidence that this improves state-tracking behaviour \citep{grazzi2025unlocking}. These recurrent architectures were chosen because they have been shown to perform well on state-tracking tasks and each has a less expressive variant for ablation: diagonal SLiCE, Mamba-2, and Gated DeltaNet with positive eigenvalues, respectively.

Each architecture is instantiated along the same width-depth schedule,
\[
d128/l1,\quad d256/l2,\quad d384/l4,\quad d512/l6,
\]
where $d$ denotes hidden dimension and $l$ denotes the number of layers. For the Transformer and Gated DeltaNet, these four settings use $2$, $4$, $6$ and $8$ heads, respectively. Mamba-3 uses the MIMO variant with rank $4$. The resulting models define a common capacity axis rather than exactly matched parameter counts, which range from $3.13$M to $3.30$M at $d128/l1$, $7.70$M to $8.67$M at $d256/l2$, $16.89$M to $20.97$M at $d384/l4$, and $35.14$M to $43.33$M at $d512/l6$. Exact parameter counts and additional implementation details for each architecture are provided in Appendix~\ref{app-model-optimisation}.

\subsection{Optimisation}

Training uses a two-stage procedure. The first stage is a one-epoch hyperparameter sweep over architecture, scale, learning rate, weight decay, warmup length, and Adam $\beta_2$. The sweep covers the four architectures and four matched scales described above. It uses batch size $128$ and AdamW optimisation, with learning rates $\{10^{-4}, 3 \times 10^{-4}, 10^{-3}, 3 \times 10^{-3}\}$, weight decays $\{0, 10^{-3}, 10^{-2}\}$, warmup steps $\{5000, 10000, 20000\}$, and Adam $\beta_2$ values $\{0.95, 0.99, 0.999\}$. This gives $1728$ one-epoch runs. The first-stage sweep uses linear warmup as the learning rate scheduler. All first-stage runs use the same seed. Validation during the sweep is performed on a fixed number of held-out real-game batches, rather than by running a full validation pass after every checkpoint. The best first-stage configuration for each family and size is reported in Appendix~\ref{app-hypopt-results}.

The second stage retrains the best-performing first-stage configurations from scratch for four epochs. These training runs use linear warmup followed by cosine decay, and tune the final learning-rate floor as a fraction of the maximum learning rate, with values in $\{0.1,0.01\}$. Appendix~\ref{app-hypopt-results} reports the second stage results, cosine-decay floor comparison, and final selected optimisation hyperparameters. This design is chosen to allow for a broad optimisation hyperparameter selection using the first stage and longer training runs of the selected configurations in the second stage.

\section{Results}
\label{sec:results}

\begin{figure}[t]
    \centering
    \includegraphics[width=0.9\textwidth]{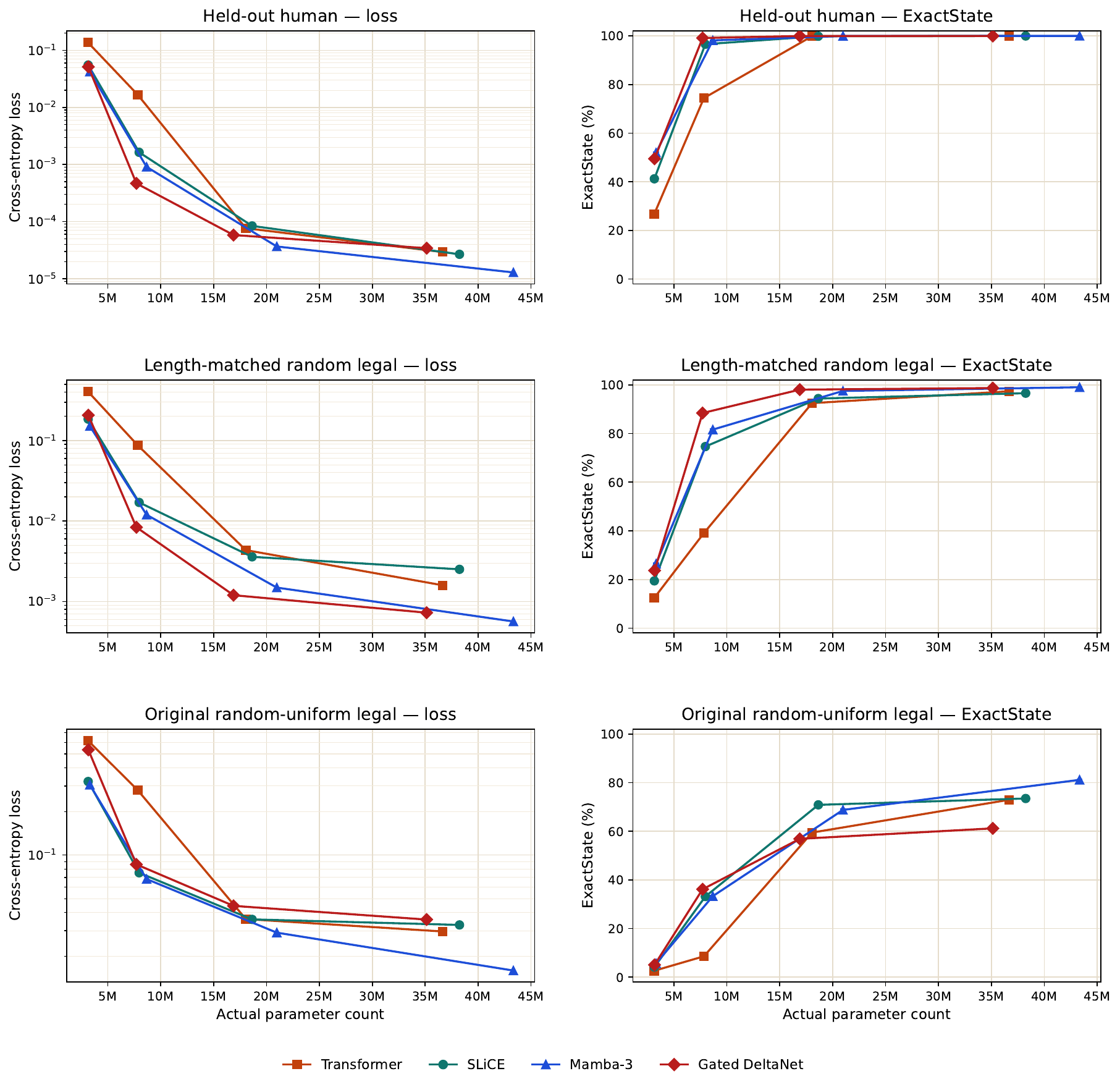}
    \caption{Test performance versus parameter count. The left panels show cross-entropy loss, and the right panels show exact state accuracy. From top to bottom, rows report held-out real-game, length-matched random, and original random performance. Lines connect configurations within each model family. Error bars show $95\%$ complete-game bootstrap confidence intervals for ExactState at fixed checkpoints.}
    \label{fig:test-metrics-vs-params}
\end{figure}

Figure~\ref{fig:test-metrics-vs-params} shows test performance as model scale increases. 
On the held-out real-game split, all recurrent models improve rapidly with scale and reach near-perfect exact state accuracy by approximately $8$M parameters. 
The Transformer is markedly less parameter-efficient at small and medium scales, obtaining substantially higher test loss and lower exact state accuracy than the recurrent models.
This suggests that chess state tracking is sensitive not only to model capacity, but also to the structure of the state-update mechanism. 
Architectures with explicit recurrent updates appear better matched to maintaining an evolving latent state from a stream of legal moves. 
However, by the largest scales, held-out real-game performance is close to saturated for all model families, and exact state accuracy no longer clearly separates the strongest recurrent configurations.

The random-uniform split gives a different picture. 
Performance continues to improve across the full scale range, and the task remains discriminative even when the held-out real-game split has largely saturated. 
The strongest model reaches around $80\%$ exact state accuracy on random-uniform games, while several models that are essentially perfect on held-out real games remain substantially below this level under random legal play. 
This indicates that high in-distribution accuracy does not imply that the model has learned robust rule-consistent state updates. 
Instead, held-out real-game evaluation can conceal failures that emerge under changes in the action distribution and trajectory lengths.
The complete held-out real-game and original random test metrics are reported in Appendix~\ref{app-test-results}.
Appendix~\ref{app-temporal-bin-results} further breaks down random-uniform performance by prefix length, showing where exact-state errors emerge along the trajectories.

The length-matched random evaluation remains discriminative after controlling the game-length distribution.
At d384/l4, human-game ExactState spans $99.85$--$99.94\%$, whereas length-matched random ExactState spans $92.51$--$98.06\%$.
At d512/l6, the corresponding ranges are $99.95$--$99.99\%$ and $96.60$--$99.03\%$.
Table~\ref{tab:matched-random-main} in Appendix~\ref{app-test-results} reports all $16$ main-model results with $95\%$ complete-game bootstrap confidence intervals.
The intervals do not overlap between architectures at any tested size, although they quantify test-sample uncertainty rather than training-seed variability.

The original random split additionally evaluates longer trajectories and can produce a different ranking.
For example, d512/l6 Gated DeltaNet is second on the matched split with $98.71\%$ ExactState, but weakest on the original random split with $61.19\%$.
These evaluations therefore expose different aspects of state tracking rather than establishing a single scale-independent architecture ranking.

\begin{figure}[t]
    \centering
    \includegraphics[width=0.9\textwidth]{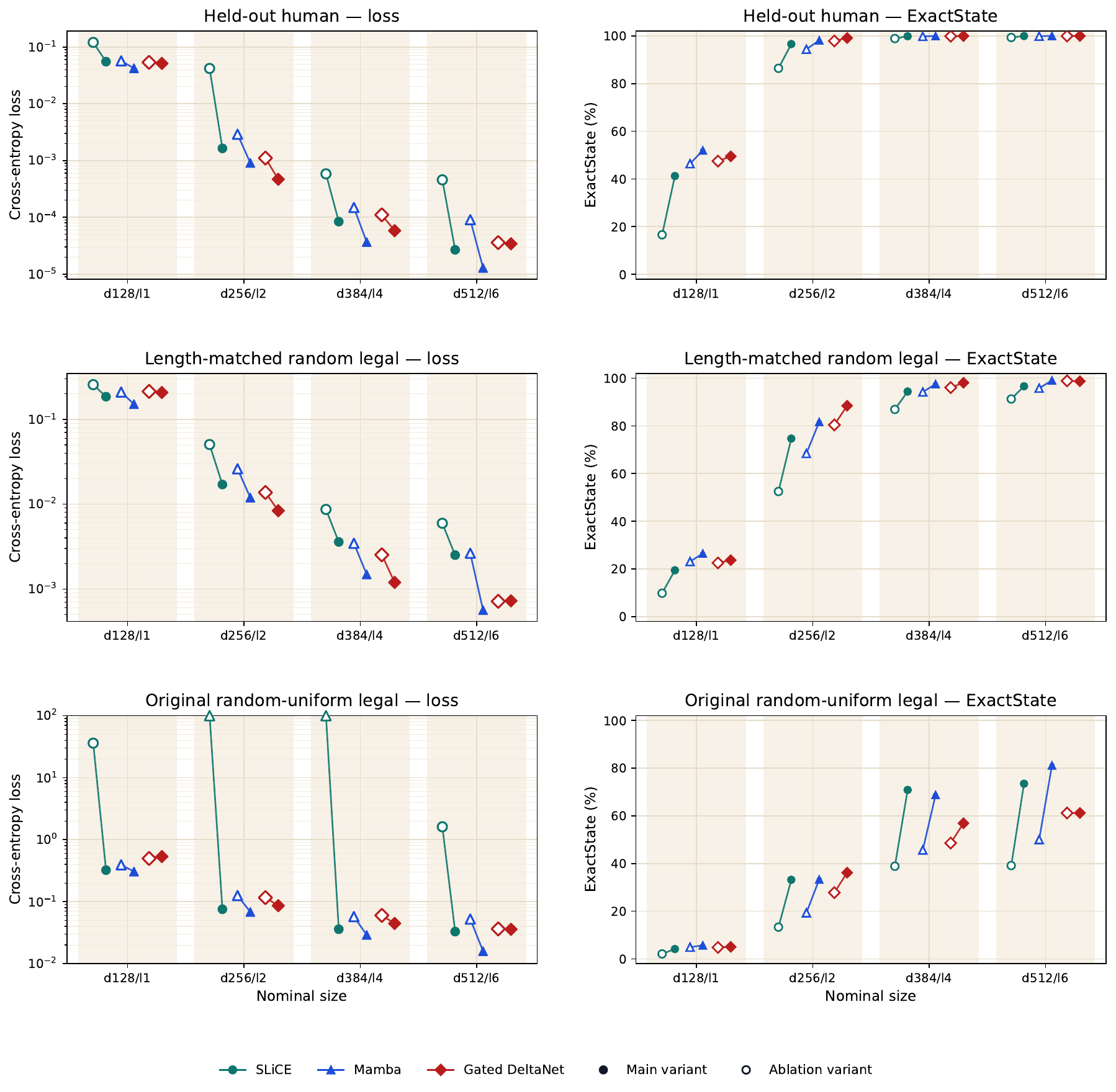}
    \caption{Comparison of recurrent state-transition variants. The left panels show cross-entropy loss, and the right panels show exact state accuracy. From top to bottom, rows report held-out real-game, length-matched random, and original random performance. Filled markers denote the main recurrent variants, while open markers denote less expressive variants. Within each shaded size group, horizontal offsets separate markers visually and do not encode an additional variable. Lines connect variants at the same nominal size. The main variants are block-diagonal SLiCE, Mamba-3, and Gated DeltaNet with negative eigenvalues; the less expressive variants are diagonal SLiCE, Mamba-2, and Gated DeltaNet with positive eigenvalues. Error bars show $95\%$ complete-game bootstrap confidence intervals for ExactState at fixed checkpoints. Upward triangles at the upper loss-axis boundary indicate off-scale values, which are reported in Appendix~\ref{app-variant-results}.}
    \label{fig:recurrent-variant-arrows}
\end{figure}

Figure~\ref{fig:recurrent-variant-arrows} compares recurrent transition mechanisms.
For each recurrent family, the less expressive variant is compared with the corresponding main variant at the same nominal size. On the held-out real-game split, the gap is most visible at smaller scales. As scale increases, the main variants and their less expressive counterparts all approach near-perfect exact state accuracy, again showing that in-distribution evaluation becomes weakly discriminative once models are sufficiently large.

On the length-matched random split, block-diagonal SLiCE improves ExactState over diagonal SLiCE by $5.34$--$22.18$ percentage points, and Mamba-3 improves over Mamba-2 by $3.23$--$13.20$ points, with all paired $95\%$ confidence intervals excluding zero. The exactly parameter-matched Gated DeltaNet comparison favours permitting negative eigenvalues through $d384/l4$, but the positive-only variant is better by $0.15$ points at $d512/l6$. The SLiCE and Mamba pairs are not parameter matched, so these results provide architecture-specific evidence rather than a fully isolated or uniform causal effect of transition expressivity. Diagonal SLiCE also exhibits numerical instability on some original random examples, producing very large cross-entropy loss. Full ablation results are given in Appendix~\ref{app-variant-results}.

Overall, these results support three conclusions. First, \textsc{Chess-World-Model} separates architecture families at small and medium scale, with recurrent models outperforming the Transformer under the matched interface. Second, held-out real-game evaluation alone is insufficient, because performance saturates and can make distinct architectures appear similarly strong. Third, the random evaluations expose failures in state tracking that are hidden by in-distribution evaluation, and the ablations generally favour more expressive recurrent transition mechanisms.

\section{Limitations and societal impact}
\label{sec:limitations}

\textsc{Chess-World-Model} isolates exact state tracking in a deterministic action domain, but it does not capture all difficulties of world modelling. Chess removes perceptual uncertainty, stochastic transitions, partial observability, continuous control, and noisy observations. Therefore, the benchmark should be viewed as a focused test of structured state updates, rather than a complete test of world modelling. 

Additionally, uniformly random legal play tests only one form of distribution shift. The original random split changes both the action distribution and the distribution of trajectory lengths. The length-matched split controls the length distribution, but conditions on random trajectories reaching the required horizon. Other agents, adversarial move sequences, longer games, or targeted tests of castling, en passant, and promotion may reveal additional weaknesses.

Furthermore, our experiments are one use of the benchmark rather than a complete evaluation of the design space. We compare several sequence-model families under a shared supervised prediction interface, so the conclusions are controlled by a common input representation, target representation, objective, and training protocol. Broader comparisons across architectures, tokenisations, objectives, scales, and training regimes are natural next steps. 
\textsc{Chess-World-Model} allows these comparisons to be made in an exact and reproducible state-tracking setting.

The direct misuse and privacy risks of \textsc{Chess-World-Model} are limited, since the benchmark uses public chess games with player metadata removed and does not release a deployed decision-making system. A more plausible risk is benchmark overclaiming: strong performance on chess state reconstruction could be misrepresented as evidence of general world-modelling ability, even though chess is fully observed, deterministic, discrete, and rule-governed. The main remaining negative impact is the compute and energy cost of large-scale benchmarking.

\section{Conclusion}
\label{sec:conclusion}

We introduced \textsc{Chess-World-Model}, a large-scale benchmark for exact state tracking from legal chess move sequences. Models predict the complete evolving chess state at every prefix of a game, including piece placement and auxiliary variables, giving an exactly checkable test of structured latent-state maintenance.

Our experiments show that recurrent models with expressive state-transition mechanisms outperform a causal Transformer at small and medium scales under a matched interface. Held-out real-game performance saturates quickly, while length-matched random evaluation remains discriminative at larger scales and the original random split additionally exposes longer-horizon failures. This shows that strong in-distribution performance can conceal failures of rule-consistent state tracking.
Ablations further show that reducing recurrent transition expressivity generally harms performance on the random-play test sets.

These results suggest that scale can conceal poor state-tracking behaviour by allowing models to learn shortcuts that perform well in distribution without capturing the underlying transition dynamics. The random-play test sets in \textsc{Chess-World-Model} exposes these failures by testing models on legal trajectories that depart from human play while preserving the same rules. This makes the benchmark a practical setting for diagnosing state-tracking failures and for developing future sequence models with more reliable internal state updates, a necessary component of more robust world models.

\newpage
\section*{Acknowledgements}

 Benjamin Walker is supported by UK Research and Innovation (UKRI) through the Engineering and Physical Sciences Research Council (EPSRC) via Programme Grant [Grant No.\ UKRI1010: High order mathematical and computational infrastructure for streamed data that enhance contemporary generative and large language models] and CIMDA@Oxford, part of the AIR@InnoHK initiative funded by the Innovation and Technology Commission, HKSAR Government.
Terry Lyons is supported by UK Research and Innovation (UKRI) through the Engineering and Physical Sciences Research Council (EPSRC) via Programme Grants [Grant No.\ UKRI1010: High order mathematical and computational infrastructure for streamed data that enhance contemporary generative and large language models], [Grant No.\ EP/S026347/1: Unparameterised multi-model data, high order signatures and the mathematics of data science], [Grant No.\ EP/Y028872/1: Mathematical Foundations of Intelligence: An Erlangen Programme for AI], and the UKRI AI for Science award [Grant No.\ UKRI2385: Creating Foundational Benchmarks for AI in Physical and Biological Complexity]. Terry Lyons is also supported by The Alan Turing Institute under the Defence and Security Programme (funded by the UK Government) and through the provision of research facilities; by the UK Government; and through CIMDA@Oxford, part of the AIR@InnoHK initiative funded by the Innovation and Technology Commission, HKSAR Government.
The authors would like to acknowledge the use of the University of Oxford Advanced Research Computing (ARC) facility in carrying out this work. https://doi.org/10.5281/zenodo.22558

\printbibliography

\newpage
\appendix

\section{Encoding details and move-token coverage}
\label{app-encoding-details}

\subsection{State-label layout}
\label{app-state-label-layout}

Table~\ref{tab-state-layout} gives the full target layout used by the benchmark.
The target contains $64$ board labels and $11$ auxiliary labels, for a total of $75$ categorical labels at each timestep.
The represented state follows the FEN-style decomposition of a chess position, including piece placement, active colour, castling availability, en passant information, the halfmove clock, and the fullmove number \citep{edwards1994pgn,fide2023laws}.
As in standard FEN, this representation does not encode the previous-position history needed to adjudicate repetition-based draws.

\begin{table}[H]
\centering
\small
\begin{tabular}{@{}p{0.34\columnwidth}p{0.16\columnwidth}p{0.38\columnwidth}@{}}
\toprule
State component & Count & Vocabulary or encoding \\
\midrule
Board squares & $64$ & $13$ labels, empty or one of $12$ piece types \\
Side to move & $1$ & white or black \\
Castling rights & $4$ & binary indicators for $K,Q,k,q$ \\
En passant file & $1$ & $9$ labels, none or files $a$--$h$ \\
En passant rank & $1$ & $3$ labels, none, rank $3$, or rank $6$ \\
Halfmove clock & $2$ & unsigned $16$-bit integer \\
Fullmove number & $2$ & unsigned $16$-bit integer \\
\bottomrule
\end{tabular}
\caption{
State representation used by the benchmark.
The target includes piece placement and the auxiliary variables of the FEN-style position state, excluding the history of previous board states needed for repetition-based draws.
}
\label{tab-state-layout}
\end{table}

\subsection{Move-token coverage}
\label{app-move-token-coverage}

The input representation packs each move into a single token determined by its source square, target square, and promotion type.
Because this vocabulary is intentionally overcomplete, it contains many source-target-promotion combinations that are invalid or unused in legal chess trajectories.
Since the random-uniform split differs from human games, a possible concern is that it might contain packed move tokens that were never observed during training.
We therefore measure identity-level move-token coverage in the training corpus and in the fixed random-uniform test set.

Let $\mathcal{V}_{\mathrm{move}}$ denote the packed non-special move vocabulary.
This vocabulary has
\[
|\mathcal{V}_{\mathrm{move}}| = 64 \times 64 \times 5 = 20{,}480
\]
possible source-target-promotion identifiers.
Table~\ref{tab-move-token-coverage} reports coverage over this non-special vocabulary.
The counts exclude the start and padding symbols.

\begin{table}[H]
\centering
\small
\begin{tabular}{lcc}
\toprule
Quantity & Train & Random-uniform test \\
\midrule
Examples kept & $10{,}000{,}000$ & $10{,}000$ \\
Unique move identifiers & $1{,}968 / 20{,}480$ & $1{,}968 / 20{,}480$ \\
Move-token coverage & $9.61\%$ & $9.61\%$ \\
\bottomrule
\end{tabular}
\caption{
Identity-level move-token coverage for the packed move vocabulary.
Although the packed vocabulary contains $20{,}480$ possible non-special move identifiers, only $1{,}968$ appear in the training set and in the fixed random-uniform test set.
}
\label{tab-move-token-coverage}
\end{table}

The observed move-token sets are identical across the two splits.
The overlap between the training set and the random-uniform test set is
\[
1{,}968 / 20{,}480 = 9.61\%,
\]
with zero random-only move identifiers and zero train-only move identifiers.
In particular, every packed move token that appears in the random-uniform test set is observed in the training data.

This rules out a simple unseen-token explanation for the random-uniform performance gap.
The random-uniform split does not test zero-shot generalisation to new packed move identities.
Instead, it tests whether models can apply the same observed move tokens in unfamiliar legal trajectories and board states.
Thus, the random-uniform split differs from the real-game training distribution in the states and action sequences it induces, not in the set of packed move-token identities it uses.

\section{Additional experimental details}
\label{app-additional-details}

\subsection{Data preprocessing, splits, and evaluation protocol}
\label{app-data-preprocessing}

Each real-game example is built by replaying a PGN game from the standard initial position using the standard chess rules \citep{edwards1994pgn,fide2023laws}.
The training and validation corpus is derived from the March 2025 Lichess open database, and the held-out real-game test set is derived from the April 2025 Lichess open database, whose public exports are released under a CC0 license \citep{lichess2026database}.
PGN parsing, legal move generation, and board-state reconstruction are implemented using \texttt{python-chess} \citep{fiekas2024pythonchess}.
The move sequence is packed into the categorical UCI-style move vocabulary, and every observed prefix is aligned with the chess state reached after replaying that prefix.
Games shorter than $10$ full moves are discarded.

The training and validation split used in the paper is deterministic at the game level. We compute the MD5 digest of each original source-game identifier, interpret its first eight bytes as an unsigned little-endian integer, and take the result modulo $10{,}000$. Games assigned to buckets below $50$ form the validation set. The public release uses opaque identifiers, so the supplied \texttt{paper\_validation\_example\_ids.txt} should be used to reproduce the paper's exact $49{,}813$-game validation set. Training shuffles the order of data shards each epoch. Validation and test use fixed datasets and evaluation order.

The random-uniform test split is generated by sampling uniformly from the legal move set at each ply until termination, treating claimable draws as terminal.
We then retain only games with at least $10$ full moves.
Final test metrics are computed on all $10{,}000$ trajectories in each of the held-out real-game, length-matched random, and original random-uniform test conditions.
During the first-stage sweep, validation is intentionally cheaper.
Models are evaluated every $5{,}000$ optimisation steps on $10$ validation batches.
With batch size $128$, this corresponds to $1{,}280$ validation games per sweep evaluation.

In addition to cross-entropy loss, we record three exactness measures.
ExactState is the fraction of timesteps for which all $75$ state labels are predicted correctly, including the $64$ board-square labels and the $11$ auxiliary labels.
Labelwise state accuracy is the average fraction of the $75$ state labels predicted correctly at a timestep.
Trajectory exactness is the fraction of complete games for which ExactState holds at every non-padding timestep.
We also record ExactState and labelwise state accuracy in contiguous $20$-step prefix bins.

\subsection{Model and optimisation details}
\label{app-model-optimisation}

All reported runs use AdamW \citep{kingma2017adam,loshchilov2019decoupled} with $\beta_1=0.9$, global gradient clipping at $1.0$ \citep{pascanu2013difficulty}, dropout $0.1$ \citep{srivastava2014dropout}, feed-forward multiplier $4$, and batch size $128$.
The value of $\beta_2$ is selected by the sweep.
On GPUs supporting tensor cores, TF32 matrix multiplication is enabled.
Mamba-3 is trained with bfloat16 mixed precision, while the other model families are trained without mixed-precision autocasting.

The first-stage sweep consists of
\[
4 \times 4 \times 4 \times 3 \times 3 \times 3 = 1728
\]
one-epoch runs over model family, model size, learning rate, warmup length, $\beta_2$, and weight decay.
The learning rates are $\{10^{-4}, 3 \times 10^{-4}, 10^{-3}, 3 \times 10^{-3}\}$.
The warmup lengths are $\{5000,10000,20000\}$.
The $\beta_2$ values are $\{0.95,0.99,0.999\}$.
The weight decays are $\{0,10^{-3},10^{-2}\}$.
These runs use linear warmup followed by an effectively constant learning rate.
All first-stage sweep runs use the same random seed.

In the second stage, selected first-stage configurations are retrained from scratch for four epochs using one of two cosine-decay schedules, with final learning rates equal to $0.1$ or $0.01$ times the maximum learning rate. Each run terminates at optimisation step $310{,}944$. The final reported configuration for each family and size is selected by held-out validation loss among the completed second-stage runs.

\paragraph{Transformer.}
The Transformer baseline is a small LLaMA-like causal decoder for autoregressive world-state prediction \citep{vaswani2017attention,touvron2023llamaopenefficientfoundation}.
It uses causal self-attention, RoPE positional embeddings \citep{su2024roformer}, RMSNorm \citep{zhang2019rmsnorm}, pre-norm residual blocks, SwiGLU MLPs \citep{shazeer2020glu}, and a final RMSNorm before the structured state-prediction heads.
The four sizes use
\[
(d,n_{\mathrm{layers}},n_{\mathrm{heads}})
=
(128,1,2),(256,2,4),(384,4,6),(512,6,8).
\]

\paragraph{SLiCE.}
The main SLiCE model uses input-dependent block-diagonal recurrent transitions \citep{walker2025slice}.
We use block size $8$, and evaluate the recurrent computation in parallel chunks of length $256$.
Each layer uses a GELU feed-forward MLP \citep{hendrycks2016gelu}, RMSNorm \citep{zhang2019rmsnorm}, and dropout on the residual path \citep{srivastava2014dropout}.
The diagonal SLiCE ablation replaces the block-diagonal transition with a diagonal transition while keeping the rest of the configuration fixed.

\paragraph{Gated DeltaNet.}
The main Gated DeltaNet model uses the same head schedule as the Transformer, together with head dimension $48$, value expansion factor $2$, gating, a short convolution of width $4$, and chunked recurrent computation \citep{yang2025gateddeltanetworksimproving}.
The main variant allows negative transition eigenvalues, following recent evidence that signed eigenvalue structure improves state tracking in linear recurrent models \citep{grazzi2025unlocking}.
The ablation restricts the transition eigenvalues to be positive while leaving the parameter count unchanged.

\paragraph{Mamba.}
The main Mamba model uses Mamba-3 with the MIMO variant and rank $4$ \citep{lahoti2026mamba3}.
It uses state dimension $128$, convolution width $4$, expansion factor $2$, head dimension $64$, one group, intermediate dimension $4d$, RMSNorm \citep{zhang2019rmsnorm}, and FP32 residual accumulation.
The Mamba-2 ablation keeps the same nominal size schedule and replaces the Mamba-3 block with Mamba-2 \citep{gu2024mamba,Dao2024Transformers}.

Table~\ref{tab-appendix-param-counts} provides the exact parameter count for each model considered in this paper.

\begin{table}[t]
\centering
\small
\begin{tabular}{@{}lrrrr@{}}
\toprule
Model & d128/l1 & d256/l2 & d384/l4 & d512/l6 \\
\midrule
Transformer & 3.13 & 7.83 & 18.05 & 36.66 \\
SLiCE & 3.14 & 7.96 & 18.64 & 38.24 \\
Mamba-3 & 3.30 & 8.67 & 20.97 & 43.33 \\
Gated DeltaNet & 3.16 & 7.70 & 16.89 & 35.14 \\
\midrule
Diagonal SLiCE & 3.03 & 7.04 & 14.50 & 27.21 \\
Mamba-2 & 3.20 & 8.23 & 19.64 & 40.65 \\
Gated DeltaNet with positive eigenvalues & 3.16 & 7.70 & 16.89 & 35.14 \\
\bottomrule
\end{tabular}
\caption{
Parameter counts in millions, rounded to two decimal places, for the evaluated configurations.
The Gated DeltaNet sign ablation leaves the parameter count unchanged.
The Mamba variants share the nominal width and depth schedule but have different parameter counts, and diagonal SLiCE is smaller than block-diagonal SLiCE.
}
\label{tab-appendix-param-counts}
\end{table}

All models were trained on a single GPU.
Training used multiple GPU types, depending on model size and kernel requirements.
The most restrictive runs were the Mamba-3 configurations, since the MIMO kernel requires an NVIDIA H100 GPU.
The longest second-stage retrains took approximately four days.
The experiments use the following publicly available datasets, libraries, and baseline implementations:

\begin{itemize}
      \item Lichess Open Database~\citep{lichess2026database}.
      License: CC0. \\
      URL: \url{https://database.lichess.org/}

      \item \texttt{python-chess}~\citep{fiekas2024pythonchess}.
      License: GPL-3.0-or-later. \\
      URL: \url{https://github.com/niklasf/python-chess}

      \item \texttt{torch-slices}~\citep{walker2025slice}.
      License: MIT. \\
      URL: \url{https://github.com/datasig-ac-uk/slices}

      \item \texttt{mamba-ssm}~\citep{gu2024mamba,Dao2024Transformers}.
      License: Apache 2.0. \\
      URL: \url{https://github.com/state-spaces/mamba}

      \item \texttt{flash-linear-attention}~\citep{yang2024fla}.
      License: MIT. \\
      URL: \url{https://github.com/fla-org/flash-linear-attention}

      \item \texttt{pytorch}~\citep{paszke2019pytorchimperativestylehighperformance}.
      License: BSD-style. \\
      URL: \url{https://github.com/pytorch/pytorch}
\end{itemize}

\subsection{Length-matched evaluation and uncertainty estimates}
\label{app-length-matched}

We construct the length-matched split from the original $10{,}000$ random test trajectories. Human-game target lengths are processed from longest to shortest. For each target length of $L$ plies, we choose the shortest unused random trajectory with at least $L$ plies and truncate it to $L$, using seeded tie ordering with seed $20260724$. The trajectories are stored in human-test order, and every original random trajectory is used once. This construction conditions on the random trajectory reaching its assigned length. The matched and original random evaluations share source trajectories and are therefore not independent.

The human and matched splits each contain $689{,}713$ plies and $699{,}713$ evaluated prefixes, with identical game-length distributions. The original random split contains $3{,}280{,}303$ plies and $3{,}290{,}303$ evaluated prefixes. Prefix counts include the initial state once per trajectory.

Metrics are pooled over valid prefixes, excluding padding. If game $i$ contributes $c_i$ exactly correct states, $n_i$ valid prefixes, and summed label cross-entropy $\ell_i$, then
\[
\mathrm{ExactState}=\frac{\sum_i c_i}{\sum_i n_i},
\qquad
\mathcal{L}=\frac{\sum_i \ell_i}{75\sum_i n_i}.
\]
For all $28$ configurations on all three splits, we compute $95\%$ confidence intervals using $10{,}000$ bootstrap resamples of complete games. Each resample draws $10{,}000$ games with replacement and recomputes the pooled metrics, preserving dependence between prefixes within each game. We use the $2.5$th and $97.5$th percentiles. Resampling indices are shared across models and metrics within each split, and paired ablation intervals use differences on the same resamples. The bootstrap seeds are $20260724$ for human and matched-random evaluation and $20260725$ for original-random evaluation. These intervals measure test-sample uncertainty for fixed checkpoints, not training-seed variability.

\section{Hyperparameter sweep and full run results}
\label{app-hypopt-results}

Table~\ref{tab-appendix-stage1-winners} reports the best one-epoch configuration for every family and size when ranked by validation loss.
The broad pattern matches the main text.
Recurrent models dominate the Transformer at small scale, and all models perform well on the in-distribution validation sweep at large scales.

Validation-loss ranking does not coincide exactly with ranking by ExactState or labelwise state accuracy.
Across the $16$ family-size cells, the validation-loss winner is also the ExactState winner in $5$ cases and the labelwise winner in $9$ cases.
In practice, the disagreements are between nearby configurations within the same family and size cell.
We therefore use validation loss as the single model-selection criterion.

\begin{table*}[t]
\centering
\small
\resizebox{\textwidth}{!}{%
\begin{tabular}{@{}llrrrrrrrr@{}}
\toprule
Model & Size & Val. loss & ExactState (\%) & Labelwise (\%) & Trajectory (\%) & LR & Warmup & $\beta_2$ & Weight decay \\
\midrule
Transformer & d128/l1 & 0.143939 & 26.02 & 95.61 & 0.16 & 0.003 & 5000 & 0.95 & 0 \\
Transformer & d256/l2 & 0.023679 & 67.15 & 99.22 & 10.62 & 0.003 & 5000 & 0.95 & 0.001 \\
Transformer & d384/l4 & 0.000959 & 98.15 & 99.97 & 77.19 & 0.003 & 5000 & 0.999 & 0.001 \\
Transformer & d512/l6 & 0.000237 & 99.51 & 99.99 & 92.19 & 0.001 & 5000 & 0.95 & 0 \\
\midrule
SLiCE & d128/l1 & 0.061826 & 38.65 & 98.03 & 0.70 & 0.001 & 20000 & 0.999 & 0.01 \\
SLiCE & d256/l2 & 0.004505 & 90.71 & 99.85 & 34.69 & 0.003 & 10000 & 0.99 & 0.001 \\
SLiCE & d384/l4 & 0.000760 & 98.45 & 99.98 & 78.44 & 0.001 & 5000 & 0.95 & 0.001 \\
SLiCE & d512/l6 & 0.000359 & 99.36 & 99.99 & 88.59 & 0.001 & 10000 & 0.95 & 0.001 \\
\midrule
Mamba-3 & d128/l1 & 0.047153 & 48.92 & 98.43 & 2.81 & 0.003 & 10000 & 0.95 & 0 \\
Mamba-3 & d256/l2 & 0.003149 & 93.58 & 99.90 & 48.28 & 0.003 & 5000 & 0.999 & 0 \\
Mamba-3 & d384/l4 & 0.000408 & 99.14 & 99.99 & 85.55 & 0.001 & 5000 & 0.95 & 0.01 \\
Mamba-3 & d512/l6 & 0.000143 & 99.70 & 100.00 & 92.81 & 0.003 & 5000 & 0.95 & 0.001 \\
\midrule
Gated DeltaNet & d128/l1 & 0.061575 & 45.51 & 98.01 & 2.42 & 0.001 & 10000 & 0.99 & 0 \\
Gated DeltaNet & d256/l2 & 0.001908 & 95.98 & 99.94 & 57.03 & 0.003 & 20000 & 0.99 & 0 \\
Gated DeltaNet & d384/l4 & 0.000448 & 99.01 & 99.99 & 84.69 & 0.001 & 5000 & 0.95 & 0 \\
Gated DeltaNet & d512/l6 & 0.000204 & 99.54 & 99.99 & 91.56 & 0.001 & 10000 & 0.95 & 0 \\
\bottomrule
\end{tabular}%
}
\caption{
Best first-stage sweep configuration in each family and size cell, ranked by validation loss on the held-out real-game split.
ExactState is the strict timestep metric requiring all $75$ state labels to be correct.
Labelwise state accuracy and trajectory exactness are measured on the same $1280$-game validation snapshot used during the sweep.
}
\label{tab-appendix-stage1-winners}
\end{table*}

The second stage retrains selected first-stage configurations for four epochs.
For every family and size except SLiCE d512/l6, the selected configuration is the first-stage validation-loss winner in Table~\ref{tab-appendix-stage1-winners}.
For SLiCE d512/l6, the first-stage validation-loss winner was unstable during the full four epoch run, so we instead chose the second-best first-stage configuration by validation loss.
For each selected configuration, we compare two cosine-decay floors, equal to $0.1$ and $0.01$ times the maximum learning rate.
Table~\ref{tab-appendix-stage2-floors} compares the two second-stage cosine-decay floors for each selected first-stage configuration.
The preferred floor depends on family and scale.
The final reported baseline in each family and size cell is the completed full run with lower held-out validation loss.

\begin{table*}[t]
\centering
\small
\resizebox{\textwidth}{!}{%
\begin{tabular}{@{}llrcrcrr@{}}
\toprule
Model & Size & Best floor & Best val. loss & Other floor & Other val. loss & ExactState (\%) & Labelwise (\%) \\
\midrule
Transformer & d128/l1 & 0.1 & 0.136054 & 0.01 & 0.136446 & 26.735 & 95.8573 \\
Transformer & d256/l2 & 0.01 & 0.015992 & 0.1 & 0.016065 & 74.984 & 99.4716 \\
Transformer & d384/l4 & 0.01 & $5.60 \times 10^{-5}$ & 0.1 & $1.14 \times 10^{-4}$ & 99.876 & 99.9982 \\
Transformer & d512/l6 & 0.1 & $7.47 \times 10^{-6}$ & 0.01 & $7.64 \times 10^{-6}$ & 99.980 & 99.9997 \\
\midrule
SLiCE & d128/l1 & 0.1 & 0.054437 & 0.01 & 0.055932 & 41.269 & 98.2439 \\
SLiCE & d256/l2 & 0.01 & 0.001462 & 0.1 & 0.001529 & 96.726 & 99.9493 \\
SLiCE & d384/l4 & 0.1 & $6.98 \times 10^{-5}$ & 0.01 & $7.28 \times 10^{-5}$ & 99.841 & 99.9977 \\
SLiCE & d512/l6 & 0.01 & $1.32 \times 10^{-5}$ & 0.1 & $1.42 \times 10^{-5}$ & 99.971 & 99.9996 \\
\midrule
Mamba-3 & d128/l1 & 0.01 & 0.041538 & 0.1 & 0.042556 & 52.194 & 98.6139 \\
Mamba-3 & d256/l2 & 0.1 & 0.000797 & 0.01 & 0.000804 & 98.216 & 99.9730 \\
Mamba-3 & d384/l4 & 0.01 & $1.18 \times 10^{-5}$ & 0.1 & $2.69 \times 10^{-5}$ & 99.982 & 99.9997 \\
Mamba-3 & d512/l6 & 0.1 & $7.27 \times 10^{-7}$ & 0.01 & $8.27 \times 10^{-6}$ & 99.999 & 100.0000 \\
\midrule
Gated DeltaNet & d128/l1 & 0.01 & 0.050420 & 0.1 & 0.050950 & 49.955 & 98.3648 \\
Gated DeltaNet & d256/l2 & 0.01 & 0.000375 & 0.1 & 0.000539 & 99.207 & 99.9888 \\
Gated DeltaNet & d384/l4 & 0.1 & $9.61 \times 10^{-6}$ & 0.01 & $1.42 \times 10^{-5}$ & 99.977 & 99.9997 \\
Gated DeltaNet & d512/l6 & 0.1 & $1.81 \times 10^{-6}$ & 0.01 & $2.43 \times 10^{-6}$ & 99.998 & 100.0000 \\
\bottomrule
\end{tabular}%
}
\caption{
Second-stage full run comparison.
Each row compares the two cosine-decay floors used in the full four-epoch runs for the selected first-stage configuration.
ExactState is the strict timestep metric requiring all $75$ state labels to be correct.
The reported ExactState and labelwise values are the validation metrics of the better full runs.
}
\label{tab-appendix-stage2-floors}
\end{table*}

Table~\ref{tab-appendix-final-hparams} consolidates the final optimisation hyperparameters used by the reported baseline models.

\begin{table*}[t]
\centering
\small
\resizebox{\textwidth}{!}{%
\begin{tabular}{@{}llrrrrr@{}}
\toprule
Model & Size & LR & Warmup & $\beta_2$ & Weight decay & LR floor \\
\midrule
Transformer & d128/l1 & 0.003 & 5000 & 0.95 & 0 & 0.1 \\
Transformer & d256/l2 & 0.003 & 5000 & 0.95 & 0.001 & 0.01 \\
Transformer & d384/l4 & 0.003 & 5000 & 0.999 & 0.001 & 0.01 \\
Transformer & d512/l6 & 0.001 & 5000 & 0.95 & 0 & 0.1 \\
\midrule
SLiCE & d128/l1 & 0.001 & 20000 & 0.999 & 0.01 & 0.1 \\
SLiCE & d256/l2 & 0.003 & 10000 & 0.99 & 0.001 & 0.01 \\
SLiCE & d384/l4 & 0.001 & 5000 & 0.95 & 0.001 & 0.1 \\
SLiCE & d512/l6 & 0.0003 & 5000 & 0.99 & 0.001 & 0.01 \\
\midrule
Mamba-3 & d128/l1 & 0.003 & 10000 & 0.95 & 0 & 0.01 \\
Mamba-3 & d256/l2 & 0.003 & 5000 & 0.999 & 0 & 0.1 \\
Mamba-3 & d384/l4 & 0.001 & 5000 & 0.95 & 0.01 & 0.01 \\
Mamba-3 & d512/l6 & 0.003 & 5000 & 0.95 & 0.001 & 0.1 \\
\midrule
Gated DeltaNet & d128/l1 & 0.001 & 10000 & 0.99 & 0 & 0.01 \\
Gated DeltaNet & d256/l2 & 0.003 & 20000 & 0.99 & 0 & 0.01 \\
Gated DeltaNet & d384/l4 & 0.001 & 5000 & 0.95 & 0 & 0.1 \\
Gated DeltaNet & d512/l6 & 0.001 & 10000 & 0.95 & 0 & 0.1 \\
\bottomrule
\end{tabular}%
}
\caption{
Final selected optimisation hyperparameters for the reported baseline models.
The LR floor is the final learning rate as a fraction of the maximum learning rate in the second-stage cosine schedule.
}
\label{tab-appendix-final-hparams}
\end{table*}

\clearpage
\section{Complete test-set results}
\label{app-test-results}

\subsection{Held-out real-game, original random, and length-matched results}

Tables~\ref{tab-appendix-test-results-real} and~\ref{tab-appendix-test-results} report held-out real-game and original random test results, respectively, for the main model families. Human-game ExactState is nearly saturated across all four families by d384/l4, while the original random split remains discriminative. Trajectory exactness remains particularly low on the original random split, indicating that few random games are tracked correctly from start to finish even at scales where human-game timestep accuracy is close to perfect. To control for the difference in game lengths, Table~\ref{tab:matched-random-main} reports all four metrics for the $16$ main-model configurations on the length-matched random split. This evaluation continues to distinguish architectures at scales where human-game accuracy is nearly saturated.

\begin{table}[htbp]
\centering
\small
\setlength{\tabcolsep}{3pt}
\begin{tabular}{@{}llcrcr@{}}
\toprule
Model & Size & Loss & \shortstack{Labelwise\\(\%)} & \shortstack{ExactState\\(\%)} & \shortstack{Trajectory\\(\%)} \\
\midrule
Transformer & d128/l1 & 0.137 [0.136, 0.139] & 95.83 & 26.69 [26.45, 26.94] & 0.02 \\
Transformer & d256/l2 & 0.0167 [0.0162, 0.0172] & 99.45 & 74.44 [74.05, 74.81] & 11.91 \\
Transformer & d384/l4 & $(7.65\,[5.52,10.7])\times10^{-5}$ & 100.00 & 99.86 [99.82, 99.89] & 97.35 \\
Transformer & d512/l6 & $(2.93\,[0.590,6.76])\times10^{-5}$ & 100.00 & 99.96 [99.92, 99.99] & 99.61 \\
\midrule
SLiCE & d128/l1 & 0.0552 [0.0546, 0.0559] & 98.22 & 41.23 [40.91, 41.55] & 0.16 \\
SLiCE & d256/l2 & $(1.64\,[1.51,1.79])\times10^{-3}$ & 99.95 & 96.57 [96.40, 96.73] & 63.95 \\
SLiCE & d384/l4 & $(8.40\,[4.99,14.1])\times10^{-5}$ & 100.00 & 99.85 [99.81, 99.89] & 97.27 \\
SLiCE & d512/l6 & $(2.67\,[1.46,4.27])\times10^{-5}$ & 100.00 & 99.95 [99.91, 99.97] & 98.97 \\
\midrule
Mamba-3 & d128/l1 & 0.0423 [0.0416, 0.0430] & 98.57 & 51.97 [51.60, 52.34] & 1.47 \\
Mamba-3 & d256/l2 & $(9.08\,[8.29,10.0])\times10^{-4}$ & 99.97 & 98.13 [98.01, 98.25] & 77.35 \\
Mamba-3 & d384/l4 & $(3.66\,[1.73,6.40])\times10^{-5}$ & 100.00 & 99.94 [99.90, 99.96] & 99.04 \\
Mamba-3 & d512/l6 & $(1.29\,[0.236,2.81])\times10^{-5}$ & 100.00 & 99.99 [99.97, 99.99] & 99.76 \\
\midrule
Gated DeltaNet & d128/l1 & 0.0514 [0.0506, 0.0523] & 98.34 & 49.47 [49.10, 49.84] & 1.03 \\
Gated DeltaNet & d256/l2 & $(4.67\,[3.79,5.93])\times10^{-4}$ & 99.99 & 99.12 [99.01, 99.22] & 87.45 \\
Gated DeltaNet & d384/l4 & $(5.81\,[1.62,13.2])\times10^{-5}$ & 100.00 & 99.93 [99.87, 99.97] & 99.46 \\
Gated DeltaNet & d512/l6 & $(3.42\,[0.542,8.78])\times10^{-5}$ & 100.00 & 99.96 [99.90, 99.99] & 99.60 \\
\bottomrule
\end{tabular}
\caption{Held-out real-game results for the $16$ main-model configurations. Brackets give $95\%$ complete-game bootstrap confidence intervals for loss and ExactState at fixed checkpoints. Labelwise accuracy and trajectory exactness are point estimates. Loss is mean cross-entropy per state label; all accuracies are percentages.}
\label{tab-appendix-test-results-real}
\end{table}

\begin{table}[htbp]
\centering
\small
\setlength{\tabcolsep}{3pt}
\begin{tabular}{@{}llcrcr@{}}
\toprule
Model & Size & Loss & \shortstack{Labelwise\\(\%)} & \shortstack{ExactState\\(\%)} & \shortstack{Trajectory\\(\%)} \\
\midrule
Transformer & d128/l1 & 0.618 [0.616, 0.620] & 85.39 & 2.63 [2.61, 2.66] & 0.00 \\
Transformer & d256/l2 & 0.281 [0.279, 0.282] & 92.48 & 8.58 [8.51, 8.65] & 0.06 \\
Transformer & d384/l4 & 0.0361 [0.0355, 0.0367] & 99.05 & 59.45 [59.12, 59.79] & 1.79 \\
Transformer & d512/l6 & 0.0296 [0.0289, 0.0303] & 99.33 & 72.93 [72.59, 73.26] & 4.87 \\
\midrule
SLiCE & d128/l1 & 0.322 [0.320, 0.325] & 92.21 & 4.14 [4.11, 4.18] & 0.00 \\
SLiCE & d256/l2 & 0.0754 [0.0738, 0.0771] & 98.19 & 33.16 [32.90, 33.42] & 0.52 \\
SLiCE & d384/l4 & 0.0359 [0.0339, 0.0380] & 99.45 & 70.85 [70.53, 71.17] & 2.54 \\
SLiCE & d512/l6 & 0.0329 [0.0312, 0.0346] & 99.49 & 73.46 [73.15, 73.78] & 4.36 \\
\midrule
Mamba-3 & d128/l1 & 0.306 [0.305, 0.307] & 91.57 & 5.66 [5.62, 5.71] & 0.00 \\
Mamba-3 & d256/l2 & 0.0682 [0.0676, 0.0689] & 98.10 & 33.31 [33.06, 33.55] & 0.90 \\
Mamba-3 & d384/l4 & 0.0291 [0.0285, 0.0297] & 99.30 & 68.78 [68.42, 69.13] & 5.06 \\
Mamba-3 & d512/l6 & 0.0159 [0.0154, 0.0165] & 99.66 & 81.13 [80.83, 81.43] & 10.83 \\
\midrule
Gated DeltaNet & d128/l1 & 0.534 [0.526, 0.543] & 90.37 & 5.05 [5.01, 5.09] & 0.00 \\
Gated DeltaNet & d256/l2 & 0.0861 [0.0850, 0.0874] & 97.90 & 36.15 [35.90, 36.40] & 1.18 \\
Gated DeltaNet & d384/l4 & 0.0445 [0.0437, 0.0454] & 99.04 & 56.88 [56.57, 57.18] & 4.52 \\
Gated DeltaNet & d512/l6 & 0.0358 [0.0351, 0.0366] & 99.22 & 61.19 [60.89, 61.49] & 5.71 \\
\bottomrule
\end{tabular}
\caption{Original random results for the $16$ main-model configurations. Brackets give $95\%$ complete-game bootstrap confidence intervals for loss and ExactState at fixed checkpoints. Labelwise accuracy and trajectory exactness are point estimates. Loss is mean cross-entropy per state label; all accuracies are percentages.}
\label{tab-appendix-test-results}
\end{table}

\begin{table}[htbp]
\centering
\small
\setlength{\tabcolsep}{3pt}
\begin{tabular}{@{}llcrcr@{}}
\toprule
Model & Size & Loss & \shortstack{Labelwise\\(\%)} & \shortstack{ExactState\\(\%)} & \shortstack{Trajectory\\(\%)} \\
\midrule
Transformer & d128/l1 & 0.407 [0.404, 0.410] & 87.56 & 12.38 [12.25, 12.51] & 0.00 \\
Transformer & d256/l2 & 0.0874 [0.0862, 0.0886] & 97.27 & 39.10 [38.75, 39.47] & 1.79 \\
Transformer & d384/l4 & $(4.35\,[4.14,4.58])\times10^{-3}$ & 99.88 & 92.51 [92.25, 92.76] & 53.81 \\
Transformer & d512/l6 & $(1.59\,[1.45,1.74])\times10^{-3}$ & 99.96 & 97.38 [97.22, 97.54] & 76.62 \\
\midrule
SLiCE & d128/l1 & 0.185 [0.183, 0.187] & 94.42 & 19.44 [19.23, 19.63] & 0.01 \\
SLiCE & d256/l2 & 0.0170 [0.0166, 0.0175] & 99.46 & 74.65 [74.21, 75.08] & 19.67 \\
SLiCE & d384/l4 & $(3.59\,[3.35,3.83])\times10^{-3}$ & 99.91 & 94.40 [94.16, 94.63] & 61.84 \\
SLiCE & d512/l6 & $(2.51\,[2.28,2.75])\times10^{-3}$ & 99.95 & 96.60 [96.41, 96.79] & 73.59 \\
\midrule
Mamba-3 & d128/l1 & 0.151 [0.150, 0.153] & 95.33 & 26.47 [26.21, 26.72] & 0.08 \\
Mamba-3 & d256/l2 & 0.0120 [0.0116, 0.0123] & 99.62 & 81.62 [81.23, 82.02] & 31.49 \\
Mamba-3 & d384/l4 & $(1.49\,[1.36,1.63])\times10^{-3}$ & 99.96 & 97.52 [97.36, 97.68] & 80.46 \\
Mamba-3 & d512/l6 & $(5.64\,[4.87,6.48])\times10^{-4}$ & 99.99 & 99.03 [98.94, 99.13] & 89.38 \\
\midrule
Gated DeltaNet & d128/l1 & 0.207 [0.205, 0.209] & 93.91 & 23.68 [23.44, 23.91] & 0.05 \\
Gated DeltaNet & d256/l2 & $(8.37\,[7.77,9.02])\times10^{-3}$ & 99.78 & 88.43 [88.12, 88.76] & 39.18 \\
Gated DeltaNet & d384/l4 & $(1.20\,[1.04,1.41])\times10^{-3}$ & 99.97 & 98.06 [97.92, 98.19] & 78.57 \\
Gated DeltaNet & d512/l6 & $(7.26\,[6.40,8.22])\times10^{-4}$ & 99.98 & 98.71 [98.61, 98.81] & 82.07 \\
\bottomrule
\end{tabular}
\caption{Length-matched random results for the $16$ main-model configurations. Brackets give $95\%$ complete-game bootstrap confidence intervals for loss and ExactState at fixed checkpoints. Labelwise accuracy and trajectory exactness are point estimates. Loss is mean cross-entropy per state label; all accuracies are percentages.}
\label{tab:matched-random-main}
\end{table}

\clearpage
\subsection{ExactState decomposition}
\label{app-state-components}

BoardExact requires all $64$ board-square labels to be correct. RulesExact requires the side to move, all four castling rights, and the en passant file and rank to be correct. No-counter Exact requires all $71$ board and rule labels to be correct. CounterExact requires both the halfmove clock and the fullmove number to be correct after reconstructing their integer values from the two predicted bytes per counter. ExactState requires all $75$ labels to be correct.

\begin{table}[htbp]
\centering
\small
\begin{tabular}{@{}lrrrrr@{}}
\toprule
Model & Board & Rules & No counters & Counters & ExactState \\
\midrule
Transformer & 81.25 & 97.53 & 81.10 & 84.85 & 72.93 \\
SLiCE & 84.97 & 98.34 & 84.64 & 84.77 & 73.46 \\
Mamba-3 & 92.08 & 99.26 & 92.01 & 86.99 & 81.13 \\
Gated DeltaNet & 87.44 & 99.93 & 87.43 & 65.92 & 61.19 \\
\bottomrule
\end{tabular}
\caption{Exact component accuracies (\%) for the d512/l6 main models on the original random split.}
\label{tab:random-decomposition}
\end{table}

Table~\ref{tab:random-decomposition} shows that errors are concentrated in board reconstruction and move counters, while rule-variable accuracy is high. Gated DeltaNet is particularly limited by the counters: its BoardExact and RulesExact are $87.44\%$ and $99.93\%$, but CounterExact is $65.92\%$. This helps explain its weaker performance on the longer original random trajectories. Component errors can co-occur, so these accuracies are not a partition of ExactState failures.

\subsection{Train--test overlap}
\label{app-train-test-overlap}

We scan $9{,}950{,}187$ training games containing $680{,}158{,}984$ states, excluding the $49{,}813$ validation games. Exact prefix identity compares the UCI move sequence from the initial position, including the empty prefix. Board-plus-rules identity compares the $64$ board labels and seven rule labels, excluding both move counters. Table~\ref{tab:train-test-overlap} reports complete-game duplication and prefix/state overlap.

\begin{table}[htbp]
\centering
\small
\setlength{\tabcolsep}{4pt}
\begin{tabular}{@{}lccc@{}}
\toprule
Split & \shortstack{Complete-game\\duplicates} & \shortstack{Exact prefix\\overlap} & \shortstack{Board-plus-rules\\overlap} \\
\midrule
Human & $5/10{,}000$ & \shortstack{$120{,}604/699{,}713$\\(17.24\%)} & \shortstack{$157{,}565/699{,}713$\\(22.52\%)} \\[2pt]
Original random & $0/10{,}000$ & \shortstack{$40{,}168/3{,}290{,}303$\\(1.22\%)} & \shortstack{$76{,}602/3{,}290{,}303$\\(2.33\%)} \\
\bottomrule
\end{tabular}
\caption{Train--test overlap. Prefix and state counts include the initial state of each game.}
\label{tab:train-test-overlap}
\end{table}

Table~\ref{tab:unseen-human} evaluates the d512/l6 models on unseen human move prefixes and unseen board-plus-rules states separately. These are distinct filtered subsets, not their intersection. ExactState remains $99.93$--$99.98\%$ under either restriction. Thus, the near-perfect human results persist when exact training prefixes or board-plus-rules states are excluded. This does not rule out all possible shortcuts or generalisation based on common human-game patterns.

\begin{table}[htbp]
\centering
\small
\begin{tabular}{@{}lrr@{}}
\toprule
Model & Unseen prefixes & Unseen board-plus-rules states \\
\midrule
Transformer & 99.95 & 99.95 \\
SLiCE & 99.93 & 99.93 \\
Mamba-3 & 99.98 & 99.98 \\
Gated DeltaNet & 99.95 & 99.95 \\
\bottomrule
\end{tabular}
\caption{Human ExactState (\%) for d512/l6 models on $579{,}109$ unseen prefixes and, separately, $542{,}148$ unseen board-plus-rules states.}
\label{tab:unseen-human}
\end{table}

\clearpage
\section{Variant results}
\label{app-variant-results}
\label{app-matched-ablations}

Tables~\ref{tab-appendix-variant-results-real}, \ref{tab-appendix-variant-results-random}, and~\ref{tab:matched-random-ablations} report held-out real-game, original random, and length-matched random results, respectively, for the $12$ less expressive recurrent variants. These ablations reduce transition expressivity while keeping the training and evaluation protocol fixed. The random evaluations are more sensitive to these changes than the held-out real-game split. On the original random split, the positive-eigenvalue Gated DeltaNet variant has lower ExactState through d384/l4 and is nearly tied at d512/l6.

Table~\ref{tab:matched-paired-ablations} reports paired ExactState differences on the length-matched split. Block-diagonal SLiCE improves over diagonal SLiCE by $5.34$--$22.18$ percentage points, and Mamba-3 improves over Mamba-2 by $3.23$--$13.20$ points, with all eight paired $95\%$ confidence intervals excluding zero. Allowing negative eigenvalues improves Gated DeltaNet through d384/l4, but the positive-only variant is better by $0.15$ points at d512/l6.

The SLiCE and Mamba pairs are not parameter matched, whereas the Gated DeltaNet pairs have identical parameter counts. Parameter count alone does not explain the SLiCE advantage: block-diagonal SLiCE d384/l4 achieves $94.40\%$ ExactState [$94.16$, $94.63$] with $18.64$M parameters, exceeding diagonal SLiCE d512/l6 at $91.26\%$ [$90.96$, $91.56$] with $27.21$M parameters. These comparisons provide architecture-specific evidence and do not isolate transition expressivity from all other architectural and optimisation differences.

\begin{table}[htbp]
\centering
\small
\setlength{\tabcolsep}{3pt}
\begin{tabular}{@{}llcrcr@{}}
\toprule
Variant & Size & Loss & \shortstack{Labelwise\\(\%)} & \shortstack{ExactState\\(\%)} & \shortstack{Trajectory\\(\%)} \\
\midrule
Diagonal SLiCE & d128/l1 & 0.121 [0.119, 0.122] & 96.48 & 16.60 [16.45, 16.75] & 0.00 \\
Diagonal SLiCE & d256/l2 & 0.0419 [0.00692, 0.112] & 99.76 & 86.40 [86.09, 86.71] & 24.98 \\
Diagonal SLiCE & d384/l4 & $(5.83\,[4.28,8.46])\times10^{-4}$ & 99.98 & 98.89 [98.76, 99.00] & 86.66 \\
Diagonal SLiCE & d512/l6 & $(4.57\,[3.63,5.94])\times10^{-4}$ & 99.99 & 99.33 [99.20, 99.43] & 92.61 \\
\midrule
Mamba-2 & d128/l1 & 0.0569 [0.0561, 0.0578] & 98.13 & 46.41 [46.05, 46.77] & 1.05 \\
Mamba-2 & d256/l2 & $(2.91\,[2.70,3.16])\times10^{-3}$ & 99.90 & 94.40 [94.18, 94.62] & 53.54 \\
Mamba-2 & d384/l4 & $(1.49\,[0.851,2.45])\times10^{-4}$ & 100.00 & 99.80 [99.73, 99.85] & 97.54 \\
Mamba-2 & d512/l6 & $(9.05\,[4.11,16.5])\times10^{-5}$ & 100.00 & 99.88 [99.81, 99.92] & 98.64 \\
\midrule
\shortstack[l]{Gated DeltaNet\\(positive)} & d128/l1 & 0.0536 [0.0528, 0.0544] & 98.26 & 47.49 [47.12, 47.85] & 0.77 \\
\shortstack[l]{Gated DeltaNet\\(positive)} & d256/l2 & $(1.10\,[0.986,1.25])\times10^{-3}$ & 99.96 & 97.87 [97.72, 98.01] & 75.99 \\
\shortstack[l]{Gated DeltaNet\\(positive)} & d384/l4 & $(1.10\,[0.580,1.81])\times10^{-4}$ & 100.00 & 99.85 [99.79, 99.90] & 98.04 \\
\shortstack[l]{Gated DeltaNet\\(positive)} & d512/l6 & $(3.58\,[1.11,7.82])\times10^{-5}$ & 100.00 & 99.95 [99.90, 99.98] & 99.55 \\
\bottomrule
\end{tabular}
\caption{Held-out real-game results for the $12$ recurrent ablations. Brackets give $95\%$ complete-game bootstrap confidence intervals for loss and ExactState at fixed checkpoints. Labelwise accuracy and trajectory exactness are point estimates. Loss is mean cross-entropy per state label; all accuracies are percentages.}
\label{tab-appendix-variant-results-real}
\end{table}

\begin{table}[htbp]
\centering
\small
\setlength{\tabcolsep}{3pt}
\begin{tabular}{@{}llcrcr@{}}
\toprule
Variant & Size & Loss & \shortstack{Labelwise\\(\%)} & \shortstack{ExactState\\(\%)} & \shortstack{Trajectory\\(\%)} \\
\midrule
Diagonal SLiCE & d128/l1 & 36.0 [29.8, 42.4] & 87.33 & 2.09 [2.07, 2.11] & 0.00 \\
Diagonal SLiCE & d256/l2 & $(4.37\,[0.00539,12.8])\times10^{17}$ & 88.96 & 13.32 [13.21, 13.43] & 0.13 \\
Diagonal SLiCE & d384/l4 & $(8.24\,[0.0348,20.3])\times10^{6}$ & 97.76 & 38.85 [38.59, 39.12] & 1.17 \\
Diagonal SLiCE & d512/l6 & 1.61 [0.540, 3.21] & 98.33 & 39.12 [38.87, 39.38] & 1.81 \\
\midrule
Mamba-2 & d128/l1 & 0.393 [0.391, 0.395] & 89.85 & 4.92 [4.88, 4.96] & 0.00 \\
Mamba-2 & d256/l2 & 0.126 [0.125, 0.127] & 96.41 & 19.32 [19.16, 19.48] & 0.44 \\
Mamba-2 & d384/l4 & 0.0575 [0.0567, 0.0582] & 98.60 & 45.67 [45.37, 45.96] & 2.49 \\
Mamba-2 & d512/l6 & 0.0524 [0.0516, 0.0533] & 98.86 & 50.02 [49.72, 50.32] & 3.04 \\
\midrule
\shortstack[l]{Gated DeltaNet\\(positive)} & d128/l1 & 0.497 [0.489, 0.505] & 90.39 & 4.80 [4.76, 4.84] & 0.00 \\
\shortstack[l]{Gated DeltaNet\\(positive)} & d256/l2 & 0.116 [0.114, 0.118] & 97.25 & 27.79 [27.58, 28.00] & 0.69 \\
\shortstack[l]{Gated DeltaNet\\(positive)} & d384/l4 & 0.0600 [0.0591, 0.0610] & 98.61 & 48.57 [48.28, 48.86] & 2.93 \\
\shortstack[l]{Gated DeltaNet\\(positive)} & d512/l6 & 0.0364 [0.0355, 0.0372] & 99.22 & 61.17 [60.87, 61.48] & 6.25 \\
\bottomrule
\end{tabular}
\caption{Original random results for the $12$ recurrent ablations. Brackets give $95\%$ complete-game bootstrap confidence intervals for loss and ExactState at fixed checkpoints. Labelwise accuracy and trajectory exactness are point estimates. Loss is mean cross-entropy per state label; all accuracies are percentages. Diagonal SLiCE exhibits numerical instability on some original random trajectories, producing very large cross-entropy loss.}
\label{tab-appendix-variant-results-random}
\end{table}

\begin{table}[htbp]
\centering
\small
\setlength{\tabcolsep}{3pt}
\begin{tabular}{@{}llcrcr@{}}
\toprule
Variant & Size & Loss & \shortstack{Labelwise\\(\%)} & \shortstack{ExactState\\(\%)} & \shortstack{Trajectory\\(\%)} \\
\midrule
Diagonal SLiCE & d128/l1 & 0.256 [0.254, 0.258] & 92.25 & 9.82 [9.73, 9.92] & 0.00 \\
Diagonal SLiCE & d256/l2 & 0.0505 [0.0447, 0.0613] & 98.58 & 52.47 [52.04, 52.90] & 5.19 \\
Diagonal SLiCE & d384/l4 & $(8.68\,[8.27,9.09])\times10^{-3}$ & 99.76 & 86.89 [86.54, 87.24] & 39.56 \\
Diagonal SLiCE & d512/l6 & $(5.96\,[5.63,6.29])\times10^{-3}$ & 99.85 & 91.26 [90.96, 91.56] & 52.09 \\
\midrule
Mamba-2 & d128/l1 & 0.209 [0.207, 0.211] & 93.65 & 23.10 [22.87, 23.33] & 0.02 \\
Mamba-2 & d256/l2 & 0.0261 [0.0255, 0.0268] & 99.16 & 68.42 [67.96, 68.87] & 16.09 \\
Mamba-2 & d384/l4 & $(3.46\,[3.27,3.66])\times10^{-3}$ & 99.90 & 94.20 [93.96, 94.44] & 63.01 \\
Mamba-2 & d512/l6 & $(2.64\,[2.47,2.83])\times10^{-3}$ & 99.93 & 95.80 [95.59, 96.00] & 66.64 \\
\midrule
\shortstack[l]{Gated DeltaNet\\(positive)} & d128/l1 & 0.213 [0.211, 0.215] & 93.77 & 22.53 [22.31, 22.75] & 0.03 \\
\shortstack[l]{Gated DeltaNet\\(positive)} & d256/l2 & 0.0137 [0.0133, 0.0142] & 99.58 & 80.38 [79.97, 80.79] & 27.34 \\
\shortstack[l]{Gated DeltaNet\\(positive)} & d384/l4 & $(2.53\,[2.32,2.77])\times10^{-3}$ & 99.94 & 96.05 [95.85, 96.25] & 68.46 \\
\shortstack[l]{Gated DeltaNet\\(positive)} & d512/l6 & $(7.17\,[6.32,8.08])\times10^{-4}$ & 99.98 & 98.87 [98.76, 98.96] & 84.93 \\
\bottomrule
\end{tabular}
\caption{Length-matched random results for the $12$ recurrent ablations. Brackets give $95\%$ complete-game bootstrap confidence intervals for loss and ExactState at fixed checkpoints. Labelwise accuracy and trajectory exactness are point estimates. Loss is mean cross-entropy per state label; all accuracies are percentages.}
\label{tab:matched-random-ablations}
\end{table}

\begin{table}[htbp]
\centering
\small
\setlength{\tabcolsep}{3pt}
\begin{tabular}{@{}lccc@{}}
\toprule
Size & SLiCE & Mamba & Gated DeltaNet \\
\midrule
d128/l1 & $+9.61\,[+9.47,+9.75]$ & $+3.37\,[+3.25,+3.49]$ & $+1.15\,[+1.06,+1.24]$ \\
d256/l2 & $+22.18\,[+21.89,+22.48]$ & $+13.20\,[+12.93,+13.49]$ & $+8.05\,[+7.78,+8.33]$ \\
d384/l4 & $+7.51\,[+7.26,+7.76]$ & $+3.33\,[+3.14,+3.51]$ & $+2.00\,[+1.85,+2.16]$ \\
d512/l6 & $+5.34\,[+5.12,+5.56]$ & $+3.23\,[+3.06,+3.41]$ & $-0.15\,[-0.25,-0.06]$ \\
\bottomrule
\end{tabular}
\caption{Paired main-minus-ablation ExactState differences on the length-matched random split, in percentage points, with $95\%$ complete-game bootstrap confidence intervals. Comparisons are block-diagonal minus diagonal SLiCE, Mamba-3 minus Mamba-2, and Gated DeltaNet allowing negative eigenvalues minus its positive-only variant. Positive values favour the main variant. Differences are computed before rounding.}
\label{tab:matched-paired-ablations}
\end{table}

\clearpage 
\section{Extended temporal-bin metrics}
\label{app-temporal-bin-results}

The timestep-averaged metrics in Table~\ref{tab-appendix-test-results} hide where errors first appear along a trajectory.
Tables~\ref{tab-appendix-bin-exactboard} and \ref{tab-appendix-bin-labelwise} therefore report saved $20$-step prefix-bin metrics on the random-uniform test set for the four d512 baseline models.

Two features stand out.
First, ExactState can collapse long before labelwise accuracy does, which means that many long-prefix errors are caused by a small number of wrong state labels.
Second, different architectures fail at different horizons.
Gated DeltaNet remains strong through roughly $180$ plies and then degrades abruptly, whereas Mamba-3 and SLiCE decay more gradually.

\begin{table*}[t]
\centering
\small
\resizebox{\textwidth}{!}{%
\begin{tabular}{@{}lrrrrrrr@{}}
\toprule
Model & 0--20 & 80--100 & 160--180 & 240--260 & 320--340 & 400--420 & 500--520 \\
\midrule
Transformer & 99.99 & 89.75 & 79.95 & 72.82 & 37.61 & 0.04 & 0.00 \\
SLiCE & 99.98 & 86.50 & 80.17 & 82.01 & 34.53 & 14.34 & 0.00 \\
Mamba-3 & 100.00 & 96.01 & 88.99 & 88.26 & 46.44 & 20.74 & 0.00 \\
Gated DeltaNet & 99.98 & 95.07 & 86.80 & 13.23 & 1.25 & 0.00 & 0.00 \\
\bottomrule
\end{tabular}%
}
\caption{
Random-uniform d512 ExactState in selected $20$-step prefix bins.
Values are percentages.
ExactState requires all $75$ state labels to be correct.
}
\label{tab-appendix-bin-exactboard}
\end{table*}

\begin{table*}[t]
\centering
\small
\resizebox{\textwidth}{!}{%
\begin{tabular}{@{}lrrrrrrr@{}}
\toprule
Model & 0--20 & 80--100 & 160--180 & 240--260 & 320--340 & 400--420 & 500--520 \\
\midrule
Transformer & 100.00 & 99.85 & 99.68 & 99.56 & 98.69 & 96.04 & 93.48 \\
SLiCE & 100.00 & 99.79 & 99.67 & 99.67 & 98.79 & 98.06 & 96.18 \\
Mamba-3 & 100.00 & 99.94 & 99.83 & 99.80 & 99.05 & 98.44 & 97.23 \\
Gated DeltaNet & 100.00 & 99.93 & 99.80 & 98.60 & 97.86 & 97.29 & 96.08 \\
\bottomrule
\end{tabular}%
}
\caption{
Random-uniform d512 labelwise state accuracy in the same selected $20$-step prefix bins as Table~\ref{tab-appendix-bin-exactboard}.
Values are percentages.
}
\label{tab-appendix-bin-labelwise}
\end{table*}

%\clearpage
%\input{checklist.tex}

\end{document}